\documentclass[journal]{IEEEtran}

\usepackage{graphicx}%
\usepackage{multirow}%
\usepackage{amsmath,amssymb,amsfonts}%
\usepackage{hyperref}
\usepackage{multirow,tabularx}
\usepackage{amsthm}%
\usepackage{mathrsfs}%
\usepackage{xcolor}%
\usepackage{textcomp}%
\usepackage{manyfoot}%
\usepackage{booktabs}%
\usepackage{algorithm}%
\usepackage{algorithmicx}%
\usepackage{algpseudocode}%
\usepackage{listings}%
\usepackage{adjustbox}

\begin{document}

\title{Leveraging Automatic Personalised Nutrition: Food Image Recognition Benchmark and Dataset based on Nutrition Taxonomy}

\author{
Sergio Romero-Tapiador*, Ruben Tolosana, Aythami Morales, Julian Fierrez, Ruben Vera-Rodriguez, Isabel Espinosa-Salinas, Gala Freixer, Enrique Carrillo de Santa Pau, Ana Ramírez de Molina and Javier Ortega-Garcia

*Corresponding author: \href{mailto:sergio.romero@uam.es}{sergio.romero@uam.es}

\thanks{S. Romero-Tapiador, R. Tolosana, A. Morales, J. Fierrez, R. Vera-Rodriguez, and J. Ortega are with the Biometrics and Data Pattern Analytics Laboratory (BiDA Lab), Escuela Politecnica Superior, Universidad Autonoma de Madrid, Spain.\\
I. Espinosa-Salinas, G. Freixer, E. Carrillo de Santa Pau and A. Ramírez de Molina are with the IMDEA Food Institute, Madrid, Spain.}
}

\maketitle
\begin{abstract}Maintaining a healthy lifestyle has become increasingly challenging in today's sedentary society marked by poor eating habits. To address this issue, both national and international organisations have made numerous efforts to promote healthier diets and increased physical activity. However, implementing these recommendations in daily life can be difficult, as they are often generic and not tailored to individuals. This study presents the AI4Food-NutritionDB database, the first nutrition database that incorporates food images and a nutrition taxonomy based on recommendations by national and international health authorities. The database offers a multi-level categorisation, comprising 6 nutritional levels, 19 main categories (e.g., ``Meat''), 73 subcategories (e.g., ``White Meat''), and 893 specific food products (e.g., ``Chicken''). The AI4Food-NutritionDB opens the doors to new food computing approaches in terms of food intake frequency, quality, and categorisation. Also,  we present a standardised experimental protocol and benchmark including three tasks based on the nutrition taxonomy (i.e., category, subcategory, and final product recognition). These resources are available to the research community, including our deep learning models trained on AI4Food-NutritionDB, which can serve as pre-trained models, achieving accurate recognition results for challenging food image databases. All these resources are available in GitHub\footnote{\url{https://github.com/BiDAlab/AI4Food-NutritionDB}}
\end{abstract}


\section{Introduction}
\IEEEPARstart{T}{he} Double Burden of Malnutrition (DBM) is defined as the coexistence of both undernutrition and overweight, a global issue affecting populations across all regions worldwide. According to the World Health Organization (WHO), it is estimated that 39\% of the adult population is currently overweight today and by 2030, this percentage will reach 50\%. In addition, Non-Communicable Diseases (NCD) have been spread in the last century mainly due to bad eating behaviours and the lack of physical activity, among others. These NCDs, such as diabetes, cardiovascular problems, or cancer, result in millions of annual deaths, emphasising the critical need for healthy dietary planning to mitigate these risks  \cite{burdenwho, obesity2030}.  

\begin{figure*}[!ht]
    \begin{center}
      \includegraphics[width=0.85\linewidth,]{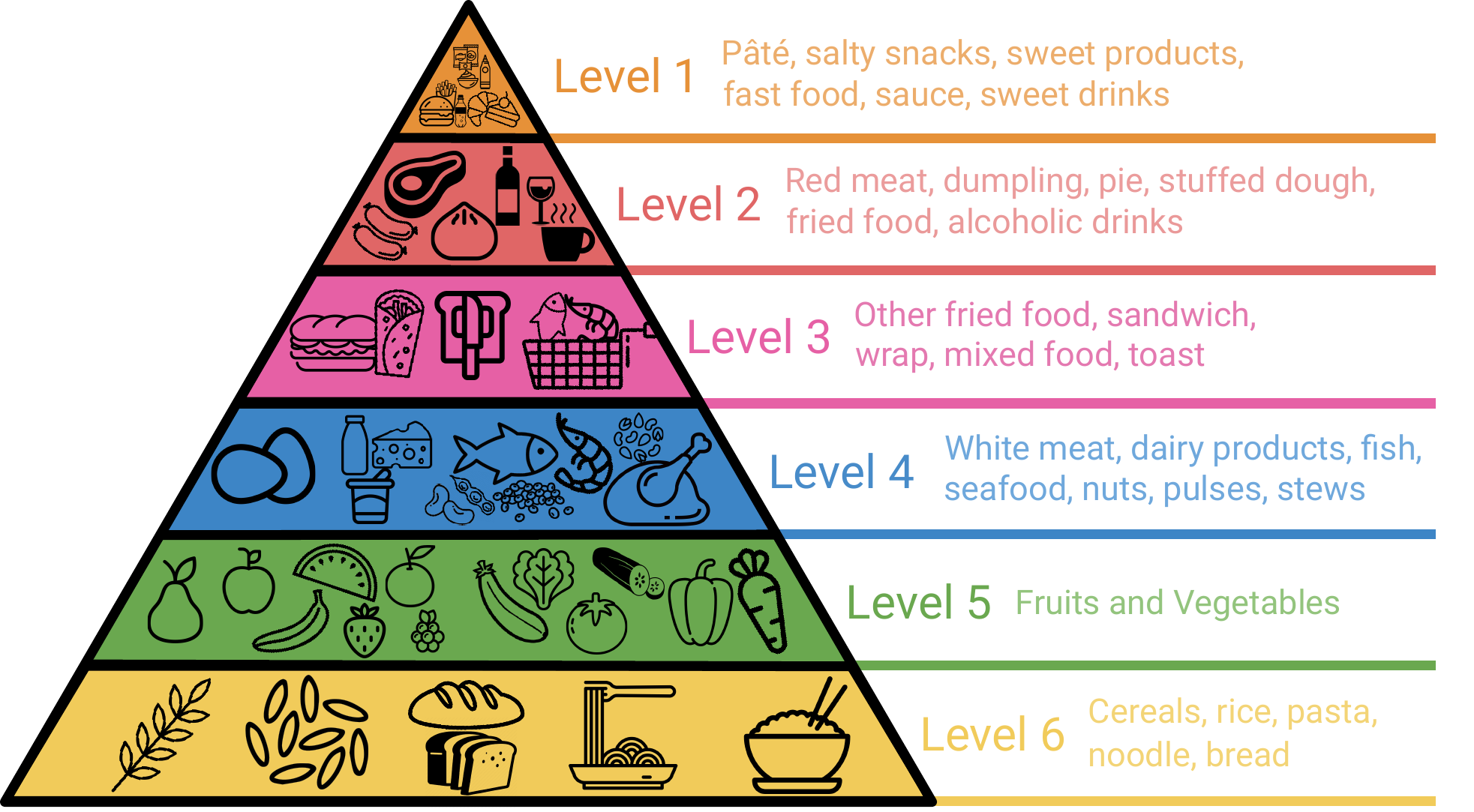}
    \end{center}
      \caption{AI4Food nutritional food pyramid. Top levels (levels 1, 2, and 3) mean lower food intake frequency, whereas bottom levels (levels 4, 5, and 6) imply higher food intake frequency.}
    \label{fig:pyramid}
\end{figure*}

Historically, strategies for promoting healthier diets have been based on recommendations tailored to the general population. National and international organisations have created food pyramids, which serve as guidelines for daily dietary choices across various food groups \cite{pyramid1, pyramid3}. Fig. \ref{fig:pyramid} provides a graphical representation of a typical food pyramid, categorising food intake into 6 nutritional levels based on intake frequency. Personalising these recommendations from general populations \cite{bida1, bida2}, together with the rapid deployment of smart devices and Artificial Intelligence (AI) methods  \cite{bida3, ml_mf}, are expected to revolutionise the promotion of healthier lifestyles.

Apart from that, health-related information, such as nutrition and physical activity, can easily be acquired through smartphones and wearable devices today \cite{bida4, smartphone_mf}. As a result, a large amount of data has been generated in recent years and their analysis, often referred to as food computing, can provide very interesting insights. Thus, food computing encompasses the acquisition and analysis of heterogeneous food data to address various food-related issues across domains such as medicine, biology, gastronomy, and agronomy \cite{foodcomputing}. For instance, a popular and user-friendly way of acquiring the eating habits of a person consists of taking pictures of the food consumed. Consequently, millions of food images have been shared through social networks and new computer vision applications based on food detection and recognition have emerged \cite{mafood, food2k, ISIA, unimib2016}. 

\begin{figure*}[!]
    \begin{center}
      \includegraphics[width=\linewidth]{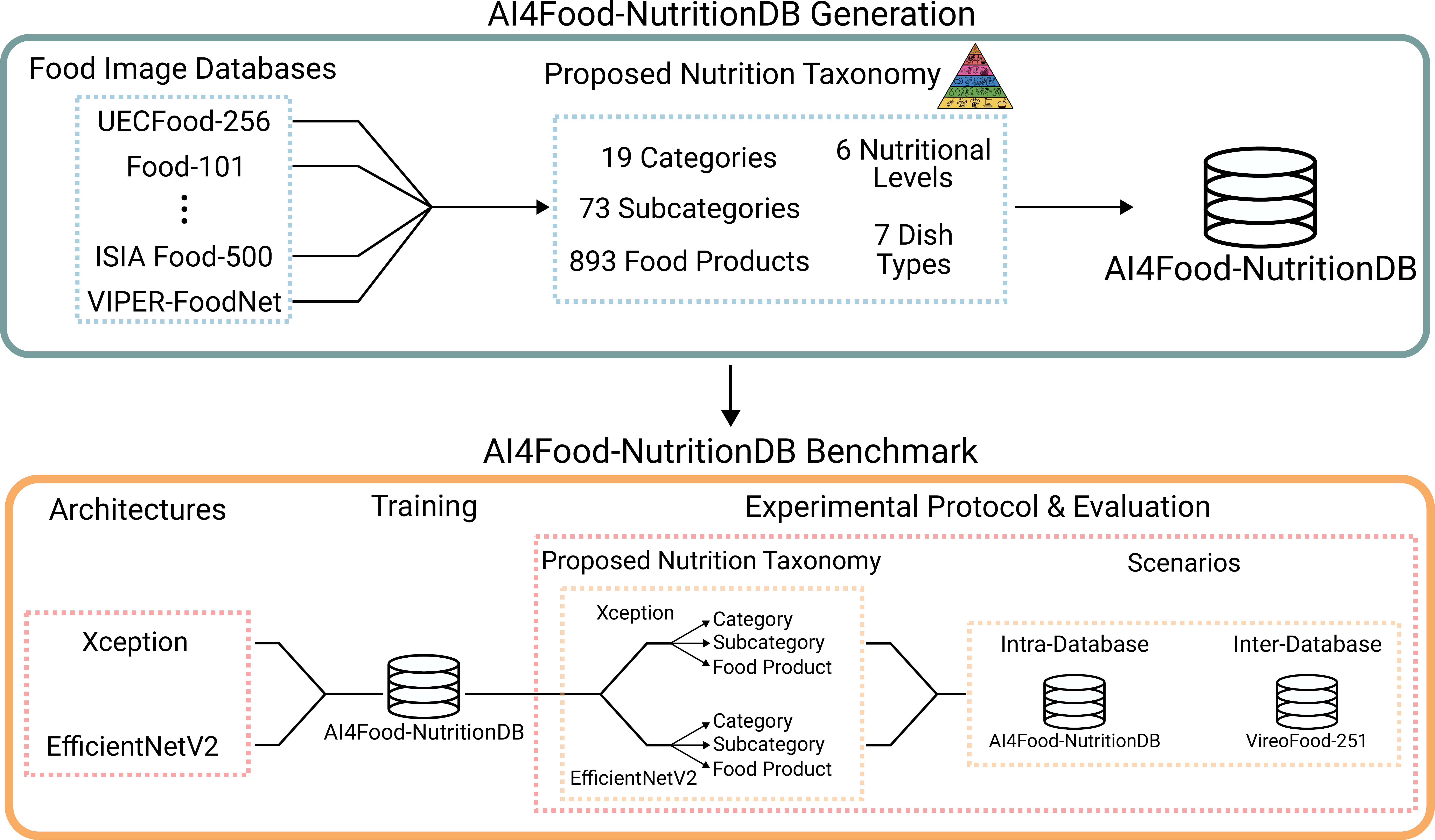}
    \end{center}
      \caption{Graphical diagram of the proposed study. First, we graphically show the AI4Food-NutritionDB database generation and the proposed nutrition taxonomy (i.e., nutritional level, category, subcategory, and food product). Second, the AI4Food-NutritionDB benchmark is presented, analysing the proposed nutrition taxonomy and scenarios (i.e., intra- and inter-database).}
    \label{fig:flow}
\end{figure*}

The current studies are limited in scope, primarily focusing on the pure application of computer vision techniques to only food images, e.g., the task of food recognition. However, the AI4Food-NutritionDB database intends to reduce the gap between resources generated by computer vision experts and the nutrition expert's guidance. First, we introduce a food image database that includes various nutrition taxonomies such as the nutritional levels of the popular food pyramid and their subcategories, among others. Second, we introduce a benchmark considering novel deep learning models to automatically assess several nutritional aspects and scenarios (i.e., intra- and inter-database). A graphical diagram of the proposed study is shown in Fig. \ref{fig:flow}.

In this article, we also present our interdisciplinary framework named AI4Food \cite{ai4fooddb}, which aims to reduce the gap between computer scientists and nutritionists. Our overall objective is to foster a new generation of technologies focused on modelling users' habits including food diet and physical activity patterns. AI4Food-NutritionDB comprises a database and a taxonomy aimed at improving current methods and resources for research in food computing. One of the primary objectives of the AI4Food framework is to create a configurable software environment capable of generating synthetic diets, including food images to simulate different user profiles depending on lifestyles and eating behaviours. This functionality has two potential applications, among many others: \textit{i)} the automatic proposal of healthy diets to the final user that can be frequently updated depending on the specific user's habits, and \textit{ii)} the automatic and continuous analysis of the eating habits of the user from the food pictures taken, giving recommendations to improve the food eating habits. To achieve this goal, this article focuses on the generation of a food nutrition image database that includes a taxonomy derived from international nutritional guidelines. 

The main contributions of the present study are:

\begin{itemize}
    \item AI4Food-NutritionDB database. To the best of our knowledge, this is the first nutrition database that considers food images and the nutrition taxonomy. The proposed taxonomy includes four different levels of categorisations, i.e., 6 nutritional levels (see Fig. \ref{fig:pyramid}), 19 main categories (e.g., the family of vertebrate animals such as ``Meat''), 73 subcategories (e.g., specific products such as ``White Meat''), and 893 final food products (e.g., final products such as ``Chicken''). In addition, each subcategory is defined by a type of dish (e.g., ``Appetizer'' and ``Main Dish'') regarding its healthiness and food quantity. Fig. \ref{fig:framework} provides a graphical description of the database and its associated taxonomy. AI4Food-NutritionDB opens the doors to new food computing approaches in terms of food intake frequency, quality, and categorisation.
    \item  Proposal of a standard experimental protocol and benchmark, including different recognition tasks (category, subcategory, and final product). The experimental protocol considered comprises both intra- and inter-database scenarios, ensuring robust evaluation. 
    \item Free release to the research community of the described datasets, protocols, and multiple deep learning models. These models can serve as pre-trained models, achieving accurate recognition results when applied to other challenging food databases. All these resources are publicly available in our GitHub repository\footnote{\url{https://github.com/BiDAlab/AI4Food-NutritionDB}}.
\end{itemize}

The remainder of the article is organised as follows. State-of-the-art studies related to food computing and food image databases are presented in Sec. \ref{sec:related}. Sec. \ref{sec:propose} explains the design of the AI4Food-NutritionDB food image database. Sec. \ref{sec:experiments} describes the proposed standard experimental protocol and benchmark results carried out on the AI4Food-NutritionDB database using deep learning techniques. Finally, conclusions and future studies are drawn up in Sec. \ref{sec:conclusion}.

\section{Related Works}\label{sec:related}

\begin{table}[!ht]
\centering
\caption{State-of-the-art food image databases. Acquisition protocol, number of classes and images, and region are described for each database. Databases considered in our AI4Food-NutritionDB database are marked in \textit{italic}. In \textbf{bold}, we highlight our food image database developed in this study. Comb. = Combination, Misc. = Miscellaneous.}
\adjustbox{width = 0.48\textwidth}{
\label{tab:ddbb}
\begin{tabular}{cccc}
\hline
\textbf{Protocol}       & \textbf{Database}                               & \textbf{\# Classes / Images}                         & \textbf{Region}                \\ \hline
                        &                                                 &                                                      &                                \\ \hline
                        & PFID (2009)                 & \multirow{2}{*}{101                    / 4,545}      & \multirow{2}{*}{Western}       \\
                        & \cite{PFID}                                     &                                                      &                                \\
                        & UNIMIB2015  (2015)           & \multirow{2}{*}{15                   / 2,000}        & \multirow{2}{*}{Misc.}         \\
                        & \cite{unimib2015}                               &                                                      &                                \\
                        & UNIMIB2016 (2016)            & \multirow{2}{*}{73                   / 1,027}        & \multirow{2}{*}{Misc.}         \\
                        & \cite{unimib2016}                               &                                                      &                                \\
                    & UNICT-FD889   (2014)         & \multirow{2}{*}{889                   / 3,583}       & \multirow{2}{*}{Misc.}         \\
          Self                  & \cite{unict889}                                 &                                                      &                                \\
           & UNICT-FD1200   (2016)       & \multirow{2}{*}{1,200                   / 4,754}     & \multirow{2}{*}{Misc.}         \\
                       & \cite{unict}                                    &                                                      &                                \\
        Collected                 & F4H            (2016)       & \multirow{2}{*}{377                     / 377}       & \multirow{2}{*}{Misc.}         \\
                        & \cite{F4H}                                      &                                                      &                                \\
                        & \textit{FruitVeg-81}   (2017)        & \multirow{2}{*}{\textit{81                     / 15,737}}     & \multirow{2}{*}{\textit{Misc.}}         \\
                        & \textit{\cite{fruitveg81}   }                            &                                                      &                                \\
                        & Mixed-Dish    (2019)                            & \multirow{2}{*}{164 / 9,254}                         & \multirow{2}{*}{Asia}          \\
                        & \cite{mixed}                                    &                                                      &                                \\
                        & Food-Pics Extended   (2019) & \multirow{2}{*}{7                    / 896}          & \multirow{2}{*}{Misc.}         \\
                        & \cite{foodpics}                                 &                                                      &                                \\ \hline
                        & Food50          (2009)                          & \multirow{2}{*}{50                    / 5,000}       & \multirow{2}{*}{Japan}         \\
                        & \cite{food50}                                   &                                                      &                                \\
                        & UECFood-100      (2012)      & \multirow{2}{*}{100                    / 14,361}     & \multirow{2}{*}{Japan}         \\
                        & \cite{UECFood100}                               &                                                      &                                \\
                        &\textit{ UECFood-256     (2014) }      & \multirow{2}{*}{\textit{256                    / 31,395}}     & \multirow{2}{*}{\textit{Japan}}         \\
                        & \textit{\cite{UECFood256} }                              &                                                      &                                \\
                        & \textit{Food-101  (2014)  }           & \multirow{2}{*}{\textit{101                    / 101,000}}    & \multirow{2}{*}{\textit{Western}}       \\
                        & \cite{Food101}                                  &                                                      &                                \\
                        & UPMCD Food-101  (2015)       & \multirow{2}{*}{101                    / 90,840}     & \multirow{2}{*}{Western}       \\
                        & \cite{upmcfood101}                              &                                                      &                                \\
                        & Dishes                          (2015)          & \multirow{2}{*}{3,832                    / 117,504}  & \multirow{2}{*}{China}         \\
                        & \cite{geodish}                                  &                                                      &                                \\
                        & Instagram 800k      (2016)                      & \multirow{2}{*}{43                   / 808,964}      & \multirow{2}{*}{Misc.}         \\
                        & \cite{instagram}                                &                                                      &                                \\
                        & VireoFood-172 (2016)        & \multirow{2}{*}{172                    / 110,241}    & \multirow{2}{*}{China}         \\
                        & \cite{VireoFood172}                             &                                                      &                                \\
                    & Food500              (2016)                     & \multirow{2}{*}{508                    / 148,408}    & \multirow{2}{*}{Misc.}         \\
                        & \cite{food500}                                  &                                                      &                                \\
      Web          & TurkishFoods-15 (2017)      & \multirow{2}{*}{15                    / 7,500}       & \multirow{2}{*}{Turkey}        \\
                         & \cite{turkish}                                  &                                                      &                                \\
                        & VegFru (2017)               & \multirow{2}{*}{292                    / 160,000}    & \multirow{2}{*}{Misc.}         \\
        Scraping                & \cite{vegfru}                                   &                                                      &                                \\
                        & ISIA Food-200 (2019)         & \multirow{2}{*}{200                    / 197,323}    & \multirow{2}{*}{Misc.}         \\
                        & \cite{isia200}                                  &                                                      &                                \\
                        & FoodX-251 (2019)            & \multirow{2}{*}{251                    / 158,846}    & \multirow{2}{*}{Misc.}         \\
                        & \cite{foodx251}                                 &                                                      &                                \\
                        & KenyanFood13 (2019)         & \multirow{2}{*}{13                    / 8,174}       & \multirow{2}{*}{Kenya}         \\
                        & \cite{kenyanfood13}                             &                                                      &                                \\
                        & Sushi50 (2019)              & \multirow{2}{*}{50                    / 3,963}       & \multirow{2}{*}{Japan}         \\
                        & \cite{sushi50}                                  &                                                      &                                \\
                        & FoodAI-756        (2019)                        & \multirow{2}{*}{756                    / 400,000}    & \multirow{2}{*}{Misc.}         \\
                        & \cite{FoodAI}                                   &                                                      &                                \\
                        & VireoFood-251   (2020)      & \multirow{2}{*}{251                    / 169,673}    & \multirow{2}{*}{China}         \\
                        & \cite{vireo251}                                 &                                                      &                                \\
                        & \textit{ISIA Food-500 (2020)    }    & \multirow{2}{*}{\textit{500                    / 399,726}}    & \multirow{2}{*}{\textit{Misc.}}         \\
                        & \textit{\cite{ISIA}}                                     &                                                      &                                \\
                        & \textit{VIPER-FoodNet (2020) }       & \multirow{2}{*}{\textit{82                     / 17,881}}     & \multirow{2}{*}{United States} \\
                        & \textit{\cite{viper}  }                                  &                                                      &                                \\
                        & Food2K    (2021)            & \multirow{2}{*}{2,000                   / 1,036,564} & \multirow{2}{*}{Misc.}         \\
                        & \cite{food2k}                                   &                                                      &                                \\ \hline
\multirow{10}{*}{Comb.} & Food201-Segmented  (2015)    & \multirow{2}{*}{201 / 50,374}                        & \multirow{2}{*}{Western}       \\
                        & \cite{food201}                                  &                                                      &                                \\
                        &\textit{ Food-11  (2016)    }         & \multirow{2}{*}{\textit{11                     / 16,643}}     & \multirow{2}{*}{\textit{Misc.}}         \\
                        & \textit{\cite{EPFL}  }                                   &                                                      &                                \\
                        & Food524DB  (2017)           & \multirow{2}{*}{524                    / 247,636}    & \multirow{2}{*}{Misc.}         \\
                        & \cite{food524db}                                &                                                      &                                \\
                        & ChineseFoodNet (2017)       & \multirow{2}{*}{208                    / 185,628}    & \multirow{2}{*}{China}         \\
                        & \cite{chinesefoodnet}                           &                                                      &                                \\
                        & \textit{MAFood-121 (2017)}           & \multirow{2}{*}{\textit{121                    / 21,175}}     & \multirow{2}{*}{\textit{Misc.}}         \\
                        & \textit{\cite{mafood} }                                  &                                                      &                                \\ \hline
\multirow{2}{*}{\textbf{Comb.}}         & \textbf{AI4Food-NutritionDB (2022) }                & \multirow{2}{*}{\textbf{893 / 558,676}}                               & \multirow{2}{*}{\textbf{Misc.}}  \\
& \textbf{Ours} & & \\ \hline
\end{tabular}}
\end{table}

\subsection{\textbf{Food Computing}}

Food computing has become a very active topic in recent years, applying computational approaches to food-related areas. Methods based on computer vision, data mining, or machine learning, among others, have been used to analyse large amounts of food images obtained from the Internet, social platforms, and smartphones. In general, food computing considers a wide range of tasks, including food segmentation, recognition, and recommendation, with applications in various fields, for instance, in health, biology, or agriculture  \cite{foodcomputing, plantSystem}. 

Among these tasks, food recognition is one of the most popular ones in the literature. This task consists of detecting and classifying food images using different techniques. Traditional approaches rely on visual features such as shape, colour, and texture for food product detection \cite{DetectorSurvey}. Scale Invariant Feature Transform (SIFT), Histogram Oriented Gradients (HOG), and Local Binary Patterns (LBP) are some popular descriptors used in the literature as feature extractors \cite{descriptor1}. For classification, Support Vector Machine (SVM) and K-Nearest Neighbour (KNN) algorithms are the most common ones to differentiate food products and categories \cite{traditional1}. However, traditional approaches are ineffective against challenging databases. In contrast, deep learning techniques have shown better performances instead \cite{food2k, results5k}. Concretely, complex architectures based on Convolutional Neural Networks (CNNs) consider both feature extraction and classification together, achieving accuracy (Acc.) rates of above 80\% \cite{unimib1, min}. For instance, Min \textit{et al.} \cite{food2k} used the architecture Squeeze-and-Excitation Network (SENet) \cite{senet}, achieving a high inter-database generalisation capacity with a 91.45\% Top-1 Acc. on the VireoFood-172 database. They also considered in \cite{food2k, ISIA} other deep learning architectures, for instance, Stacked Global-Local Attention Network (SGLANet) and Progressive Region Enhancement Network (PRENet). Experiments were carried out using the ISIA Food-500 and Food2K databases \cite{ISIA, food2k}, achieving Top-1 Acc. results of 64.74\% and 83.62\%, respectively. 

It is important to highlight that most published studies focus on food recognition at the final food product level (using as labels the name of the dish, for example, ``Pasta alla Norma'') or the main category (e.g., ``Fast Food"). However, in the present article, we analyse the task of food recognition based on the proposed nutrition taxonomy (6 nutritional levels, 19 main categories, 73 subcategories, and 893 final food products) as this is needed for many real applications, particularly those related to healthy dietary practices. In addition, each subcategory is defined by a type of dish (e.g., ``Appetizer'' and ``Main Dish'') regarding its healthiness and food quantity.

\subsection{\textbf{Food Image Databases}}

Many food image databases have emerged in recent years, including a wide range of food products from various world areas. These databases are categorized based on three different acquisition protocols found in the literature (i.e., self-collected, web scraping, and combination). Table \ref{tab:ddbb} provides a summary of these databases and their metadata, including the protocol used, the number of classes and images, and the world region:

\begin{itemize}
    \item \textbf{Self-collected}: this consists of taking food images from a camera or smartphone in controlled or semi-controlled environments. Although databases such as PFID \cite{PFID}, UNIMIB2015 \cite{unimib2015}, or UNIMIB2016 \cite{unimib2016} include a large variety of food products, the number of total images is relatively low ($<$ 20K images) due to the extensive manual process. Similarly, UNICT-FD889 \cite{unict889} and UNICT-FD1200 \cite{unict} are two databases with 889 and 1,200 final food products, respectively, which represent food dishes from different parts of the world and nationalities (e.g., English, Japanese, Indian, and Italian, among others). In addition, FruitVeg-81 \cite{fruitveg81} contains more than 15,000 fruit and vegetable images, and Mixed-Dish \cite{mixed}, in contrast, is a food image database of 164 Asian food products. Finally, F4H \cite{F4H} and Food-Pics Extended \cite{foodpics} are two databases captured in a controlled scenario and a plain background. 
    
    \item \textbf{Web Scraping}: web scraping techniques are employed to acquire large amounts of food images. In contrast to self-collected techniques, thousands of food images can be easily captured from social and web platforms. Some databases focus on food products from specific regions of the world, for instance, traditional Japanese and Chinese dishes (e.g., Food50 \cite{food50}, UECFood-256 \cite{UECFood256}, VireoFood-251 \cite{vireo251})  while others include dishes from Europe and North America (e.g., Food-101 \cite{Food101}, UPMCD Food-101 \cite{upmcfood101}). Additionally, databases like TurkishFoods-15 \cite{turkish}, KenyanFood13 \cite{kenyanfood13}, and VIPER-FoodNet \cite{viper} are three food image databases from Turkey, Kenya, and the United States, respectively. Other databases include food dishes from several regions of the world. Instagram 800k \cite{instagram}, Food500 \cite{food500}, ISIA Food-200 \cite{isia200}, and FoodX-251 \cite{foodx251}. Similarly to FruitVeg-81, VegFru \cite{vegfru} contains only fruit and vegetable images. ISIA Food-500 \cite{ISIA} is a database with 500 food final food products and lastly, Food2K \cite{food2k} is a recent database with around 1M food images organised into 2K food products.    
    \item \textbf{Combination}: this consists in creating new food image databases by combining data from existing ones. For instance, Food201-Segmented \cite{food201}, is derived from the Food-101 database, supplemented with food tags using crowd-sourcing. Food-11 \cite{EPFL} is a database created primarily from three different databases (Food-101, UECFOOD-100, and UECFOOD-256) and grouped into 11 main food categories.  Multi-Attribute Food-121 (MAFood-121) \cite{mafood} database comprises the top-11 most popular cuisines in the world according to Google Trends (such as French, Mexican, or Vietnamese cuisines) and comprises 121 final food products and more than 21K food images.     
\end{itemize}

To summarise, various food image databases have been presented in the literature considering different acquisition approaches and conditions. However, none of them have previously incorporated a nutrition taxonomy that assesses the quality, quantity, and intake frequency of foods based on images. The database proposed in this study offers a nutritional categorisation that facilitates the development of a new generation of food computing algorithms that foster its use in various food-related areas.

\section{AI4Food-NutritionDB Database}\label{sec:propose}
\begin{figure*}[t]
    \begin{center}
      \includegraphics[width=\linewidth]{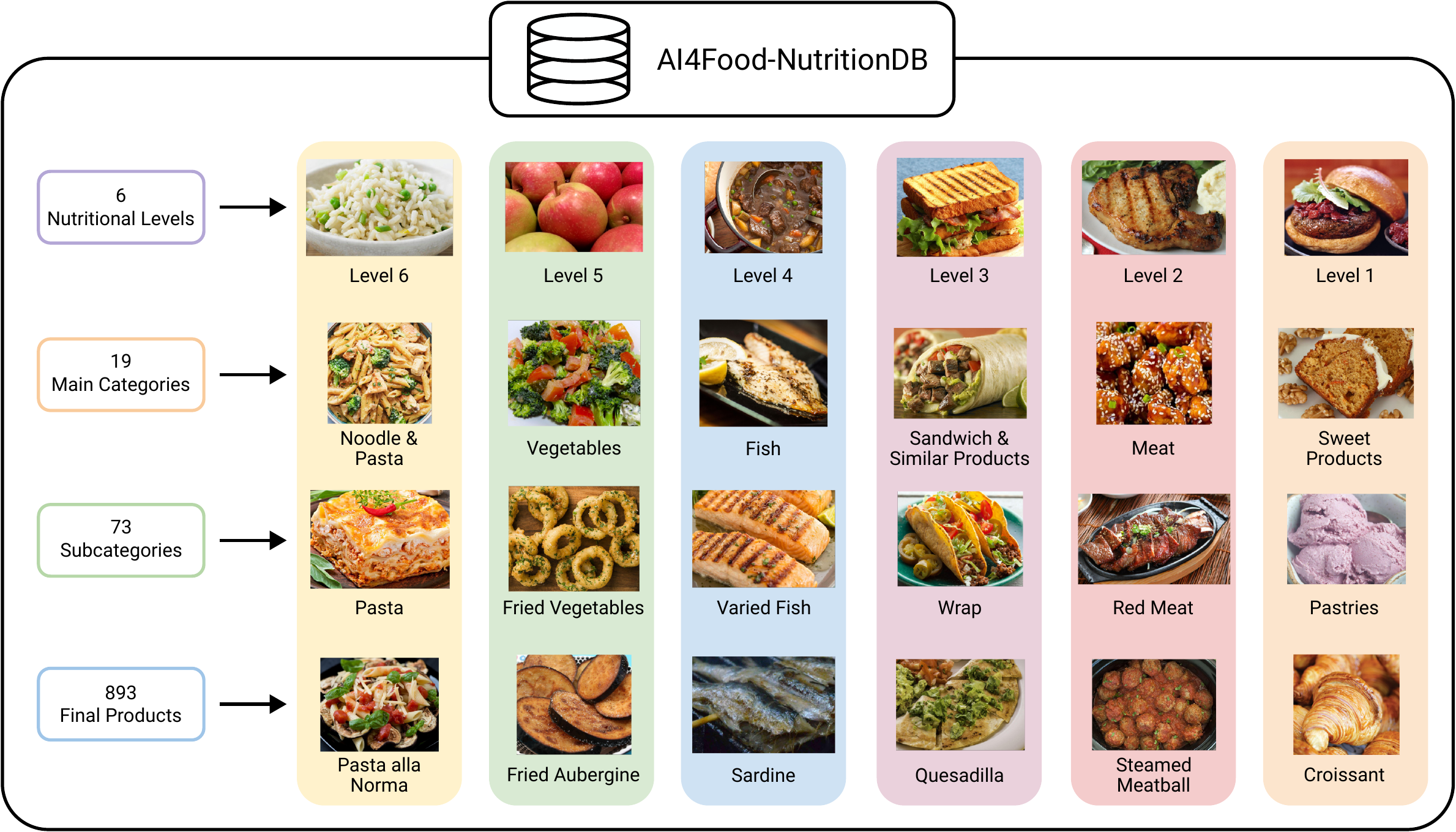}
    \end{center}
    \caption{Description of the AI4Food-NutritionDB food image database framework and taxonomy. This database is generated using food images from seven different state-of-the-art databases, i.e., UECFood-256 \cite{UECFood256}, Food-101 \cite{Food101}, Food-11 \cite{EPFL}, FruitVeg-81 \cite{fruitveg81}, MAFood-121 \cite{mafood}, ISIA Food-500 \cite{ISIA}, and VIPER-FoodNet \cite{viper}. AI4Food-NutritionDB database comprises 6 nutritional levels, 19 main categories, 73 subcategories, and 893 final products with over 500K food images. }
    \label{fig:framework}
\end{figure*}

The proposed AI4Food-NutritionDB is the first nutrition database that considers food images and a nutrition taxonomy. This taxonomy includes four different levels of categorisation: 6 nutritional levels (see Fig. \ref{fig:pyramid}), 19 main categories (e.g., the family of vertebrate animals such as ``Meat''), 73 subcategories (e.g., specific products such as ``White Meat''), and 893 final food products (e.g., final products such as ``Chicken''). In addition, each subcategory is defined by a type of dish (e.g., ``Appetizer'' and ``Main Dish'') considering factors related to healthiness and food quantity. Fig. \ref{fig:framework} provides a graphical description of the database. AI4Food-NutritionDB has been built by combining food images from seven different databases, encompassing food products from all over the world. We provide next all the information regarding the source databases (Sec. \ref{sub:3a}) and the construction process of the AI4Food-NutritionDB (Sec. \ref{sub:3b}).

\subsection{\textbf{Source Food Image Databases}}\label{sub:3a}

Seven state-of-the-art food image databases were selected to construct our database. These databases encompass various world regions and exhibit different characteristics. 

\subsubsection[UECFood-256]{\textbf{UECFood-256}{\footnote{\url{http://foodcam.mobi/dataset256.html}}} \cite{UECFood256}}
UECFood-256 contains 256 food products and more than 30K Japanese food images from different platforms such as Bing Image Search, Flickr, and Twitter (web scraping acquisition). In addition, they employed Amazon Mechanical Turk (AMT) for image selection and labelling.

\subsubsection[Food-101]{\textbf{Food-101}\footnote{\url{https://data.vision.ee.ethz.ch/cvl/datasets\_extra/food-101/}}  \cite{Food101}}
This database comprises over 100K food images and 101 unique food products from various world regions. All the images were sourced from the FoodSpotting application, a social platform where individuals uploaded and shared food images. 

\subsubsection[Food-11]{\textbf{Food-11}\footnote{\url{https://www.kaggle.com/vermaavi/food11}} \cite{EPFL}}
Singla \textit{et al.} analysed the eating behaviour to construct a database that comprised some of the food groups consumed in the United States. This way they defined 11 general food categories from the  United States Department of Agriculture (USDA), including bread, dairy products, dessert, eggs, fried food, meat, noodle/pasta, rice, seafood, soups, and vegetables/fruits. They also combined three different databases (Food-101, UECFood-100, and UECFood-256) and two social platforms (Flickr and Instagram) to finally accumulate more than 16K food images.

\subsubsection[FruitVeg-81]{\textbf{FruitVeg-81}\footnote{\url{https://www.tugraz.at/institute/icg/research/team-bischof/lrs/downloads/fruitveg81/}} \cite{fruitveg81}}
Many of the state-of-the-art food image databases do not consider many fruit or vegetable food products. As a distinctive feature, this database contains images in these mentioned groups highly underrepresented. FruitVeg-81 database has 81 different fruits and vegetable food products acquired from the self-collected acquisition protocol.

\subsubsection[MAFood-121]{\textbf{MAFood-121}\footnote{\url{http://www.ub.edu/cvub/mafood121/}} \cite{mafood}} 
Considering the 11 most popular cuisines in the world (according to Google Trends), Aguilar \textit{et al.} released the MAFood-121 database. This database contains 121 unique food products and around 21K food images grouped into 10 main categories (bread, eggs, fried food, meat, noodle/pasta, rice, seafood, soup, dumpling, and vegetables). They utilised the combination acquisition protocol, using three state-of-the-art public databases (Food-101, UECFood-256, and TurkishFoods-15) and a private one.

\begin{table}[t]

\renewcommand{\thetable}{\Roman{table}A}
         \caption{Description of the categories, subcategories, number of products, nutritional levels, and types of dish considered in the AI4Food-NutritionDB database. }
\parbox{\linewidth}{
        \vspace{\baselineskip}
                          
        \centering

        \adjustbox{width=0.49\textwidth}{
            \begin{tabular}{cccccc}

                \hline
\textbf{Category} & \textbf{Subcategory} & \textbf{Products}   & \textbf{Nutr. Level} & \textbf{Dish}           \\ \hline
Rice                            & Rice                 & 14 & 6 & Main    \\ \hline
                                & Noodle               & 57 & 6 & Main    \\
Noodle \&                       & Italian Pasta        & 13 & 6 & Main    \\
Pasta             & Other Types          & \multirow{2}{*}{4}  & \multirow{2}{*}{6}   & \multirow{2}{*}{Main}   \\
                                & of Pasta             &    &   &         \\ \hline
Bread                           & Bread                & 4  & 6 & Bread   \\
\&                              & Toast                & 6  & 3 & Dessert \\
Similar           & Other Types          & \multirow{2}{*}{5}  & \multirow{2}{*}{3}   & \multirow{2}{*}{Appet.} \\
Products                        & of Bread             &    &   &         \\ \hline
\multirow{4}{*}{Vegetables}     & Fresh Vegetables     & 20 & 5 & Main    \\
                                & Mushrooms            & 2  & 5 & Main    \\
                                & Cooked Vegetables    & 14 & 5 & Main    \\
                                & Fried Vegetables     & 11 & 4 & Main    \\ \hline
Fruits                          & Fruits               & 29 & 5 & Dessert \\ \hline
                                & Side Dish Salad      & 3  & 5 & Side    \\
Salad             & Other Types          & \multirow{2}{*}{19} & \multirow{2}{*}{3}   & \multirow{2}{*}{Main}   \\
                                & of Salad             &    &   &         \\ \hline
\multirow{6}{*}{Meat}           & White Meat           & 14 & 4 & Main    \\
                                & Red Meat             & 26 & 2 & Main    \\
                                & Breaded Meat         & 18 & 2 & Main    \\
                                & Varied Meat          & 21 & 2 & Main    \\
                                & Sausage              & 2  & 2 & Main    \\
                                & Pâté                 & 1  & 1 & Appet.  \\ \hline
\multirow{3}{*}{Fish}           & Varied Fish          & 20 & 4 & Main    \\
                  & Fried or             & \multirow{2}{*}{7}  & \multirow{2}{*}{3}   & \multirow{2}{*}{Main}   \\
                                & Breaded Fish         &    &   &         \\ \hline
\multirow{4}{*}{Seafood}        & Mollusk              & 6  & 4 & Main    \\
                                & Crustacean           & 5  & 4 & Main    \\
                                & Fried Seafood        & 3  & 3 & Main    \\
                                & Varied Seafood       & 1  & 4 & Main    \\ \hline
\multirow{3}{*}{Beans}          & Fresh Beans          & 1  & 4 & Main    \\
                                & Cooked Beans         & 10 & 4 & Main    \\
                                & Fried Beans          & 2  & 3 & Main    \\ \hline
Eggs                            & Eggs                 & 6  & 4 & Main    \\ \hline
                                & Cheese               & 1  & 4 & Appet.  \\
Dairy                           & Yogurt               & 1  & 4 & Appet.  \\
Products          & Fried Dairy          & \multirow{2}{*}{2}  & \multirow{2}{*}{3}   & \multirow{2}{*}{Main}   \\
                                & Products             &    &   &         \\ \hline
Soups \&                        & Soups and Creams     & 62 & 4 & Main    \\
Stews                           & Stews                & 21 & 4 & Main    \\ \hline
Sandwich \&                     & Sandwich             & 21 & 3 & Main    \\
Similar Products                & Wrap                 & 7  & 3 & Main    \\ \hline
\multirow{4}{*}{Salty Snacks}   & Nut Snacks           & 2  & 4 & Snack   \\
                                & Vegetable Snacks     & 3  & 1 & Snack   \\
                                & Bean Snacks          & 3  & 1 & Snack   \\
                                & Other Salty Snacks   & 2  & 1 & Snack   \\ \hline
\multirow{5}{*}{Sweet Products} & Pastries             & 89 & 1 & Dessert \\
                                & Chocolate Products   & 7  & 1 & Dessert \\
                                & Dairy Dessert        & 14 & 1 & Dessert \\
                                & Fruit Dessert        & 10 & 1 & Dessert \\
                                & Other Sweet Products & 16 & 1 & Dessert \\ \hline

            \end{tabular}
    }}
     \label{tab:cat}

\end{table}

\begin{table}[t]
\addtocounter{table}{-1}
\renewcommand{\thetable}{\Roman{table}B}
        \caption{Continuing from Table \ref{tab:cat}.}
\parbox{\linewidth}{
        \vspace{\baselineskip}
        \centering

        \adjustbox{width=0.49\textwidth}{
            \begin{tabular}{cccccc}
                \hline
\textbf{Category}       & \textbf{Subcategory} & \textbf{Products} & \textbf{Nutr. Level} & \textbf{Dish} \\ \hline
                                & Pizza                & 6  & 1 & Main    \\
Fast                            & Burger               & 3  & 1 & Main    \\
Food                            & Hot Dog              & 3  & 1 & Main    \\
                                & Fries                & 3  & 1 & Side    \\ \hline
      & Mixed Rice          & 23 & 6 & Main   \\
\multicolumn{1}{l}{}    & Mixed Vegetables     & 16                & 5                    & Main          \\
      & Rice and Fish       & 2  & 4 & Main   \\
      & Rice and Beans      & 4  & 4 & Main   \\
      & Mixed Fish          & 9  & 4 & Main   \\
      & Mixed Seafood       & 14 & 4 & Main   \\
      & Mixed Beans         & 7  & 4 & Main   \\
      & Mixed Eggs          & 8  & 4 & Main   \\
Mixed & Sushi               & 5  & 4 & Main   \\
      & Rice and Meat       & 15 & 3 & Main   \\
Food  & Meat and Vegetables & 20 & 3 & Main   \\
      & Mixed Food          & 53 & 3 & Main   \\
      & Mixed Meat          & 48 & 2 & Main   \\
      & Dumpling            & 10 & 2 & Main   \\
      & Pie                 & 3  & 2 & Main   \\
      & Stuffed Dough       & 4  & 2 & Main   \\
      & Fried Food          & 14 & 2 & Main   \\
      & Sauce               & 5  & 1 & Snack  \\ \hline
\multirow{5}{*}{Drinks} & Vegetable Drinks     & 2                 & 3                    & Drinks        \\
      & Coffee              & 2  & 2 & Drinks \\
      & Alcoholic Drinks    & 3  & 2 & Drinks \\
      & Sugary Drinks       & 3  & 1 & Drinks \\
      & Other Drinks        & 3  & 1 & Drinks \\ \hline
            \end{tabular}
        }}
       \label{tab:cat2}
\end{table}

\subsubsection[ISIA]{\textbf{ISIA Food-500}\footnote{\url{http://123.57.42.89/FoodComputing-Dataset/ISIA-Food500.html}} \cite{ISIA}}
ISIA Food-500 is a database released in 2020. All food images (around 400K) are organised into 500 different food products and were acquired from Google, Baidu, and Bing search engines, including both Western and Eastern cuisines. Following a similar approach to other databases, they categorised all food products into 11 major groups, including meat, cereal, vegetables, seafood, fruits, dairy products, bakery products, fat products, pastries, drinks, and eggs.

\subsubsection[VIPER-FoodNet]{\textbf{VIPER-FoodNet}\footnote{\url{https://lorenz.ecn.purdue.edu/~vfn/}} \cite{viper}}
Similar to the Food-11 database, VIPER-FoodNet is an 82-food-product database, selected based on the most commonly consumed items in the United States from the What We Eat In America (WWEIA) database\footnote{\url{https://data.nal.usda.gov/dataset/what-we-eat-america-wweia-database}}. All the images were obtained through web scraping, specifically from Google Images.

As a result, the proposed AI4Food-NutritionDB initially comprises 1,152 food products with 586,914 food images. This diverse database represents traditional dishes from several world areas, such as Food-101 with Western dishes, UECFood-256 with traditional Japanese dishes, and VIPER FoodNet, with typical food dishes from the United States. In addition, the ISIA Food-500 database has 500 food products from various countries, and the FruitVeg-81 database also includes fruit and vegetable images from several world regions. Finally, it is important to highlight that Food-11 and MAFood-121 databases are created from some of the previous databases. As a result, post-processing was conducted to remove duplicated images.

\begin{figure*}[!]
    \begin{center}
      \includegraphics[width=\linewidth]{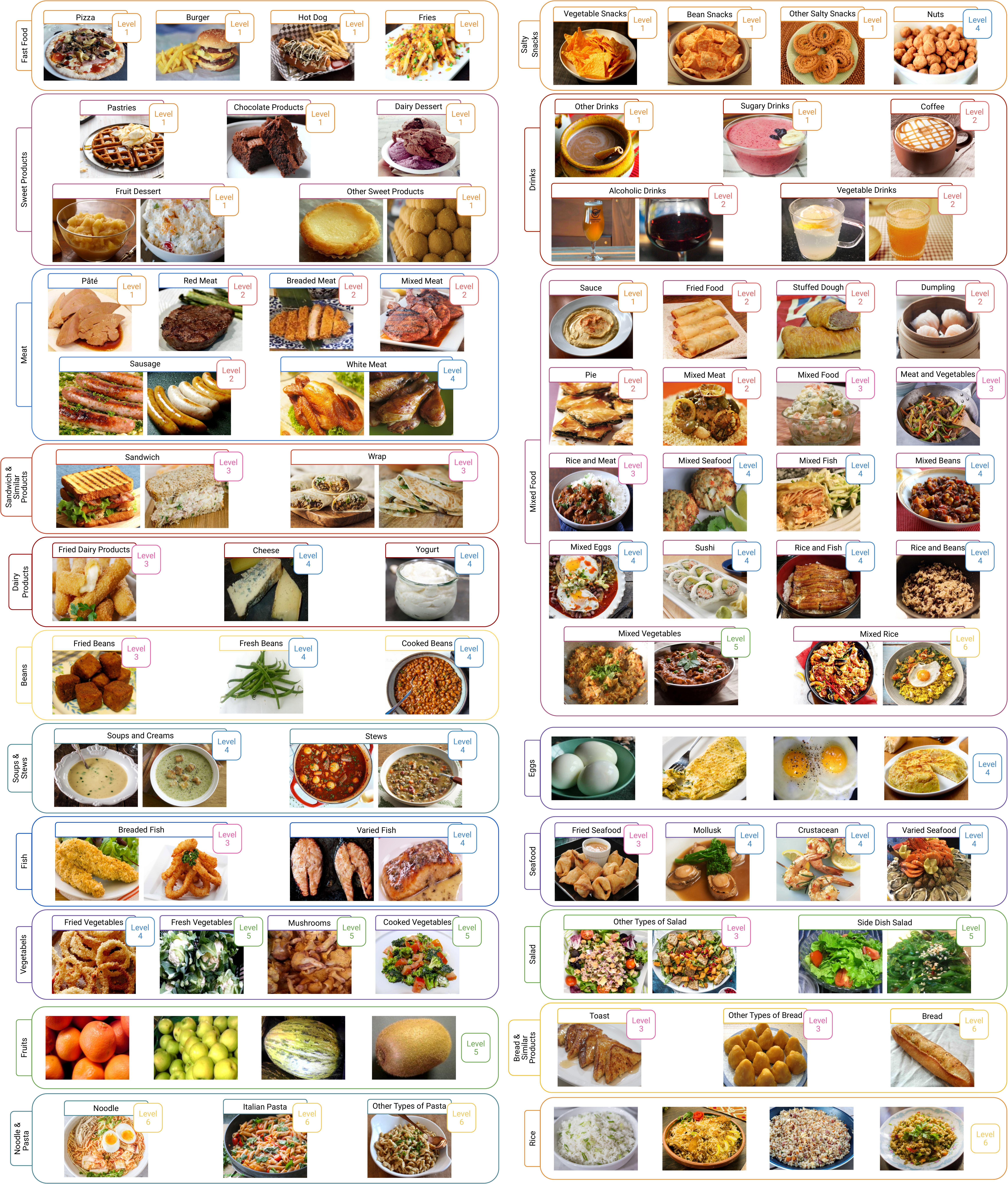}
    \end{center}
      \caption{Graphical representation of the categories and subcategories within the AI4Food-NutritionDB food image database. Note that the placement of the categories has been attempted to align with the pyramid in Fig. \ref{fig:pyramid}. As can be observed, this positioning generates ambiguities and discrepancies (e.g., for mixed and cooked food) that were resolved as described in Sec. \ref{sub:3b}.}
    \label{fig:ddbb}
\end{figure*}

\subsection{\textbf{Food Product Categorisation}}\label{sub:3b}
Each of the 1,152 food products obtained in the previous stage is individually processed for classification into the following levels: \textit{i)} nutritional level, \textit{ii)} category, \textit{iii)} subcategory, and \textit{iv)} type of dish. Tables \ref{tab:cat} and \ref{tab:cat2} summarise information from the AI4Food-NutritionDB database, including the levels, the number of products, and the type of dish for each subcategory. For completeness, we provide in Fig. \ref{fig:ddbb} a graphical representation of the categories, subcategories, and nutritional levels considered in AI4Food-NutritionDB. Each subcategory features one or two food images labelled with its corresponding nutritional level, and subcategories are further grouped into main categories.
    
\subsubsection{\textbf{Process of Categorisation}}
Three different stages are considered to classify each food product into subcategories, categories, and types of dishes. In the initial stage, the food product's taxonomy is extracted using FoodOn ontology\footnote{\url{https://www.ebi.ac.uk/ols/ontologies/foodon}} \cite{foodon}, which provides supercategories and subcategories for corresponding food products. Although this ontology comprises over 9K food products, a high percentage of the analysed items is not contemplated by FoodOn, particularly those translated from their original language (e.g., the Danish plate \textit{\ae blefl\ae sk}). The second stage involves querying the food term within the TasteAtlas web platform\footnote{\url{https://www.tasteatlas.com/}}, which contains around 10K traditional dishes worldwide. In addition, metadata such as ingredients, dish type, or food region is included in this platform. 

Finally, in the third stage, each final food product is classified into subcategories and categories, and all examined food terms go through a review and unification process. This step involves eliminating terms that do not comply with established criteria and merging those with similar characteristics. The outcome of this process yields a collection of 893 final food products.

\subsubsection{\textbf{Nutritional Level}}
The nutritional level indicates the intake frequency for a specific food product and is determined by the popular nutritional pyramids proposed by national and international organisations, such as the United States Department of Agriculture (USDA) pyramid \cite{pyramid1} and the Spanish Society of Community Nutrition (SENC) pyramid \cite{pyramid3}. Fig. \ref{fig:pyramid} provides a graphical representation of the typical food pyramid considered for AI4Food-NutritionDB based on 6 different nutritional levels. A lower nutritional level (at the pyramid's top) implies limited consumption, whereas a higher level (at the pyramid's bottom), denotes greater intake. 

Most food products align with the different nutritional levels proposed in the pyramid, allowing the nutritional level assignment. However, some food products or subcategories are not directly contemplated by nutritional level, e.g., ``Fried Vegetables'' or ``Rice and Fish''. For all these ambiguous cases, the nutrition experts of the AI4Food framework have manually defined the appropriate nutritional level.

\subsubsection{\textbf{Dish Type}} 
Following a similar process to the nutritional level assignment, each subcategory is set to a dish type to differentiate it from others that can be found during a meal, since the quantity of each dish significantly varies. Seven different types of dishes are defined, following the guidelines established in \cite{typedish}: 

\begin{itemize}
    \item \textbf{Main Dish}: this type of dish represents most of the subcategories defined and includes both first and second courses.
    \item \textbf{Appetizer}: this dish is usually consumed before the main dish and the quantity is relatively less. ``Pâté'', ``Cheese'', and ``Other Types of Bread'' are included in it.
    \item \textbf{Snack}: similar to an appetizer, this is consumed at any time of the day. All the ``Salty Snack'' subcategories and ``Sauce'' comprise this dish.
    \item \textbf{Dessert}: usually consumed at the end of a meal, dessert often consists of sweet food products. In this study, the ``Fruits'' and ``Toast'' subcategories, and ``Sweet Products'' category, are included in the Dessert dish type.
    \item \textbf{Side Dish}: served with main dishes such as ``Fries'' and ``Side Dish Salad''.
    \item \textbf{Bread}: this basic food product is usually eaten alongside main dishes. In this case, only the ``Bread'' subcategory is included.
    \item \textbf{Drinks}: this is represented by the ``Drinks'' products.
\end{itemize}

As a result, the AI4Food-NutritionDB database comprises 558,676 food images grouped into 6 nutritional levels, 19 main categories, 73 subcategories, and 893 food products as depicted in Tables \ref{tab:cat} and \ref{tab:cat2}.

\section{AI4Food-NutritionDB Benchmark}\label{sec:experiments}

This section describes the proposal of a standard experimental protocol and benchmark evaluation of AI4Food-NutritionDB, based on the nutrition taxonomy (category, subcategory, and final product). First, the deep learning recognition systems are described in Sec. \ref{subsec:4a}. Then, Sec. \ref{subsec:4b} describes the proposed experimental protocol. Finally, Sec. \ref{subsec:4c} and Sec. \ref{subsec:4d} provide the recognition results achieved on intra- and inter-database scenarios, respectively. 

In addition, we provide the complete experimental protocol, benchmark evaluation, and pre-trained models, all of which are available on our GitHub repository\footnote{\url{https://github.com/BiDAlab/AI4Food-NutritionDB}}. The repository contains detailed documentation and instructions for reproducing our experiments and scenarios.

\begin{table*}[t]
\centering
\caption{\textbf{Intra-database results in terms of Top-1 and Top-5 Acc. over the AI4Food-NutritionDB database.} Three different scenarios and two state-of-the-art deep learning architectures are considered based on the nutrition taxonomy, i.e., category (19 classes), subcategory (73 classes), and final products (893 classes). Each model is trained with its corresponding number of classes and is evaluated using the final test subsets. The best results achieved are marked in \textbf{bold} for each dataset and nutrition taxonomy (category, subcategory, and product).}
\label{tab:resultsintra}
\resizebox{\textwidth}{!}{%
\begin{tabular}{l|cccc|cccc|cccc|}
\cmidrule{2-13}
\multicolumn{1}{c|}{\multirow{3}{*}{\textbf{\textbf{Database}}}} & \multicolumn{4}{c|}{\textbf{Category Performance}} & \multicolumn{4}{c|}{\textbf{Subcategory Performance}} & \multicolumn{4}{c|}{\textbf{Product Performance}} \\\cmidrule{2-13} 
 & \multicolumn{2}{c|}{Xception} & \multicolumn{2}{c|}{EfficientNetV2} & \multicolumn{2}{c|}{Xception} & \multicolumn{2}{c|}{EfficientNetV2} & \multicolumn{2}{c|}{Xception} & \multicolumn{2}{c|}{EfficientNetV2} \\ \cmidrule{2-13}
 & Top-1 & \multicolumn{1}{c|}{Top-5} & Top-1 & Top-5 & Top-1 & \multicolumn{1}{c|}{Top-5} & Top-1 & Top-5 & Top-1 & \multicolumn{1}{c|}{Top-5} & Top-1 & Top-5 \\ \hline
 \textbf{- AI4Food-NutritionDB (ours)} & \multicolumn{1}{c}{77.74} & \multicolumn{1}{c|}{97.78} & \multicolumn{1}{c}{\textbf{82.04}} & \textbf{98.45} & 70.62 & \multicolumn{1}{c|}{92.01} & \textbf{77.66} & \textbf{95.40} & 55.60 & \multicolumn{1}{c|}{81.39} & \textbf{66.28} & \textbf{88.82} \\ 
\quad - UECFood-256 \cite{UECFood256} & {77.81} & \multicolumn{1}{c|}{{98.54}} & \textbf{83.19} & \textbf{98.97} & 68.13 & \multicolumn{1}{c|}{91.94} & \textbf{79.26} & \textbf{96.78} & 61.08 & \multicolumn{1}{c|}{84.19} & \textbf{76.89} &  \textbf{94.36}\\
\quad - Food-101    \cite{Food101} & {82.59} & \multicolumn{1}{c|}{{97.99}} & \textbf{87.63} & \textbf{98.88} & 76.86 & \multicolumn{1}{c|}{93.79} & \textbf{84.15} & \textbf{97.05} & 70.22 & \multicolumn{1}{c|}{89.80} & \textbf{84.56} & \textbf{96.20} \\
\quad - Food-11      \cite{EPFL} & {97.30} & \multicolumn{1}{c|}{\textbf{{100}}} & \multicolumn{1}{c}{\textbf{100}} & \multicolumn{1}{c|}{\textbf{100}}  & \textbf{70.27} & \multicolumn{1}{c|}{{\textbf{100}}} & 56.76 & \multicolumn{1}{c|}{\textbf{100}} & - & \multicolumn{1}{c|}{-} & \multicolumn{1}{c}{-} & \multicolumn{1}{c|}{-} \\
\quad - FruitVeg-81   \cite{fruitveg81} & {\textbf{99.63}} & \multicolumn{1}{c|}{{\textbf{100}}} & 98.73 & 99.97 & \textbf{99.46} & \multicolumn{1}{c|}{99.97} & 99.36 & \multicolumn{1}{c|}{\textbf{100}} & 98.88 & \multicolumn{1}{c|}{{\textbf{100}}} & \textbf{99.49} & 99.90 \\
\quad - MAFood-121   \cite{mafood} & {71.62} & \multicolumn{1}{c|}{{97.88}} & \multicolumn{1}{c}{{\textbf{82.25}}} & \textbf{99.55} & 61.15 & \multicolumn{1}{c|}{87.71} & \textbf{74.20} & \textbf{94.23} & 44.05 & \multicolumn{1}{c|}{67.84} & \textbf{64.49} & \textbf{88.88} \\
\quad - ISIA Food-500   \cite{ISIA} & {76.54} & \multicolumn{1}{c|}{{97.79}} & \textbf{78.29} & \textbf{98.25} & 69.19 & \multicolumn{1}{c|}{91.69} &  \textbf{76.06}& \textbf{96.32} & 51.77 & \multicolumn{1}{c|}{79.55} & \textbf{61.46} & \textbf{86.94}    \\
\quad - VIPER-FoodNet    \cite{viper} & {61.51} & \multicolumn{1}{c|}{{90.65}} & \textbf{65.47} & \textbf{93.12} & 50.96 & \multicolumn{1}{c|}{81.14} &\multicolumn{1}{c}{\textbf{58.55}} & \textbf{87.77} & 32.79 & \multicolumn{1}{c|}{60.81} & \textbf{50.56} & \textbf{78.43} \\ \hline

\end{tabular}%
}
\end{table*}

\subsection{Proposed Food Recognition Systems}\label{subsec:4a}

The proposed food recognition systems utilize state-of-the-art CNN architectures, namely Xception \cite{xception} and EfficientNetV2\footnote{\url{https://github.com/sebastian-sz/efficientnet-v2-keras}} \cite{efficientnetv2}. These architectures have been selected due to outstanding performances in computer vision tasks such as food recognition,  deepfake detection, and image classification in general ~\cite{tolosanaromero, chen2021review, morales2023exploring}. First, the Xception approach is inspired by Inception \cite{inception}, replacing Inception modules with depthwise separable convolutions. Secondly, the EfficientNetV2 approach is an optimised model within the EfficientNet family of architectures, able to achieve better results with fewer parameters compared to other models in challenging databases like ImageNet \cite{imagenet}. 

In this study, we follow the same training approach considered in \cite{tolosanaromero},  using a pre-trained model with ImageNet, where the last fully-connected layers are replaced with the number of classes specific to each experiment. Then, all the weights from the model are fixed up to the fully-connected layers and re-trained for over 10 epochs. Subsequently, the entire network is trained again for 50 more epochs, choosing the best-performing model in terms of validation accuracy. We use the following features for all experiments, employing an Adam optimiser based on binary cross-entropy using a learning rate of $10^-3$, and $\beta_1$ and $\beta_2$ of $0.9$ and $0.999$, respectively. In addition, training and testing are performed with an image size of 224$\times$224. The experimental protocol was executed with the aid of an NVIDIA GeForce RTX 4090 GPU, utilising the Keras library.

\subsection{Experimental Protocol}\label{subsec:4b}
For reproducibility reasons, we adopt the same experimental protocol considered in the collected databases,  dividing them into development and test subsets following each corresponding subdivision. In addition, the development subset is also divided into train and validation subsets. However, three of the collected databases -FruitVeg81, UECFood-256, and Food-101- do not contain this division. In such cases, we employ a similar procedure as presented in \cite{tolosanaromero}. Around 80\% of the images comprise the development subset, with the train and validation subsets also distributed around 80\% and 20\% of the development subset, respectively. The remaining images correspond to the test subset (around 20\%). It is important to remark that no images are duplicated across the three subsets (train, validation, and test) in any of the seven databases. Similarly to \cite{food2k}, Top-1 (Top-1 Acc.) and Top-5 classification accuracy (Top-5 Acc.) are used as evaluation metrics. 

\begin{table}[t]
\centering

\caption{\textbf{Inter-database results in terms of Top-1 and Top-5 Acc. over the VireoFood-251 database.} Two different deep learning models (Xception and EfficientNetV2) are considered, both pre-trained with ImageNet and AI4Food-NutritionDB databases. The best results achieved are remarked in \textbf{bold} for each architecture.}
\label{tab:inter}
\adjustbox{width = 0.49\textwidth}{
\begin{tabular}{ccccc}
\hline
\textbf{\begin{tabular}[c]{@{}c@{}}Deep Learning\\ Model\end{tabular}} & \textbf{Fine-tuned Database} & \textbf{\begin{tabular}[c]{@{}c@{}}Training \\ Strategy\end{tabular}} & \textbf{Top-1} & \textbf{Top-5} \\ \hline
\multirow{4}{*}{Xception \cite{xception}} & ImageNet & - & 58.91 & 83.78 \\
 & \multirow{3}{*}{AI4Food-NutritionDB} & Category & 81.38 & 95.37 \\
 &  & Subcategory & 81.78 & 95.56 \\
 &  & Product & \textbf{82.10} & \textbf{95.71} \\ \hline
\multirow{4}{*}{EfficientNetV2 \cite{efficientnetv2}} & ImageNet & - &  81.54& 96.25 \\
 & \multirow{3}{*}{AI4Food-NutritionDB} & Category & 88.22 & 97.95 \\
 &  & Subcategory & \textbf{88.80} & \textbf{98.07} \\
 &  & Product & 88.70 & 98.02 \\ \hline
\end{tabular}
}
\end{table}

\subsection{Intra-database Results}\label{subsec:4c}
Three different scenarios are considered for the intra-database evaluation of the AI4Food-NutritionDB database. Each scenario represents a different level of granularity defined by the number of categories (19), subcategories (73), and final products (893). Table \ref{tab:resultsintra} summarises the performances obtained for the different intra-database scenarios and deep learning architectures considered in the AI4Food-NutritionDB. For completeness, we also include the results achieved on the individual databases included in AI4Food-NutritionDB. We highlight the best results in \textbf{bold} for each dataset and nutrition taxonomy. This allows us to also assess the model's performance across the different subsets. Regarding the whole AI4Food-NutritionDB database, category scenario performances show the best results, obtaining 77.74\% Top-1 Acc. and 97.78\% Top-5 Acc. for Xception, and 82.04\% Top-1 Acc. and 98.45\% Top-5 Acc. for EfficientNetV2. However, the performance significantly drops as the granularity becomes finer for both architectures. For example, for the EfficientNetV2 architecture, the Top-1 Acc. decreases from 82.04\% to 77.66\% and 66.28\% for the subcategory (73 classes) and product (893 classes) analysis, respectively. This decrease is mainly due to the similarity in appearance among different subcategories (e.g., ``White Meat'' and ``Red Meat''),  final products (e.g., ``Pizza Carbonara'' and ``Pizza Pugliese''), or even the same food cooked in several manners (e.g., ``Baked Salmon'' and ``Cured Salmon''). Regarding each specific dataset, the FruitVeg-81 dataset shows the best results in general for both deep learning architectures, classifying almost perfectly the different fruits and vegetables (over 98\% Top-1 and Top-5 Acc. for all categorisation scenarios). Contrarily, the VIPER-FoodNet dataset obtains the worst results in each categorisation scenario as images sometimes contain food products with mixed ingredients (e.g., different types of beans, meat, and pasta). Finally, in terms of the deep learning architecture, EfficientNetV2 outperforms Xception in all scenarios (category, subcategory, product) of the AI4Food-NutritionDB for both Top-1 Acc. and Top-5 Acc. metrics. These results highlight the potential of the state-of-the-art EfficientNetV2 architecture for the nutrition taxonomy proposed in the present article.

\subsection{Inter-database Results}\label{subsec:4d}
To assess the generalisation ability of our deep learning models pre-trained with the proposed AI4Food-NutritionDB, we include an inter-database scenario using the challenging VireoFood-251 food image database \cite{vireo251}, which is an extended version of VireoFood-172 \cite{VireoFood172}. This database comprises over 169K food images distributed in 251 Chinese food plates, which were not included in AI4Food-NutritionDB. In this experiment, we consider two different scenarios based on the training process. First, we consider XceptionNet and EfficientNetV2 architectures both pre-trained only with ImageNet \cite{imagenet}. Secondly, we consider again both architectures but pre-trained with AI4Food-NutritionDB. In the last case, three different models are considered, each trained at a different level of granularity following our proposed nutrition taxonomy (category, subcategory, and final product). In order to reproduce the same experimental protocol proposed by the authors, we only train the last fully-connected layers from each pre-trained model for 30 epochs, freezing the rest of the model. Table \ref{tab:inter} shows the final test results obtained in each scenario for the final product categorisation (i.e., 251 Chinese food plates). Again, the best performances are marked in \textbf{bold} for each deep learning model. The results indicate that using our pre-trained models with AI4Food-NutritionDB improve the performance in terms of both Top-1 and Top-5 Acc. in comparison with the models pre-trained with only the ImageNet database. For instance, for the Xception architecture, the model pre-trained with the AI4Food-NutritionDB achieves for the final product categorisation results of 82.10\% Top-1 and 95.71\% Top-5 Acc., much better results in comparison with the 58.91\% Top-1 and 83.78\% Top-5 Acc. obtained with the ImageNet model. For the EfficientNetV2 architecture, results are even better with 88.80\% Top-1 Acc. and 98.07\% Top-5 Acc. Therefore, the proposed deep learning models trained with the proposed AI4Food-NutritionDB can effectively serve as reliable pre-trained models, achieving accurate recognition results with unseen food databases.

\section{Conclusion and Future Study}\label{sec:conclusion}

This article presents the AI4Food-NutritionDB, the first database with a nutrition taxonomy and over 560K food images. Furthermore, we propose a standardised experimental protocol and benchmark for the AI4Food-NutritionDB, utilising food recognition systems based on two state-of-the-art architectures. Our evaluation encompasses both intra- and inter-database scenarios across different food recognition levels. We finally prove that our pre-trained models using AI4Food-NutritionDB can improve state-of-the-art food recognition systems in challenging scenarios. Our contribution facilitates the development of novel food computing approaches that foster a better understanding of what we eat.

This study opens several future research lines including the improvement of the database by incorporating new taxonomy levels from nutritional experts (e.g., based on the nutritional composition of the ingredients or the composition of the prepared food at hand). On the other hand, behavioural habits (e.g., physical activity, sleep quality) are key factors strongly related to the impact of nutrition on our health \cite{romerotapiador2023ai4foodnutritionfw}. Future studies will significantly benefit by incorporating comprehensive multimodal models of user habits towards personalised interventions adapted to individual characteristics and necessities \cite{ai4fooddb}.  For instance, new studies could focus on the impact of glucose from food intake on metabolic health or the impact of sleep quality on dietary patterns \cite{fontana2021detection}. We also plan to integrate statistical \cite{bida5} and human-readable food descriptors through recent Large Language Models (LLMs) \cite{10445251} into our framework to improve both classification rates and interpretability of our models \cite{bida6}.

\section*{Funding}

This study has been supported by projects: AI4FOOD-CM (Y2020/TCS6654), FACINGLCOVID-CM (PD2022-004-REACT-EU), INTER-ACTION (PID2021-126521OB-I00 MICINN/FEDER), and HumanCAIC (TED2021-131787BI00 MICINN).


\end{document}